**Review** **Open Access**

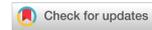

# Intelligent feature extraction, data fusion and detection of concrete bridge cracks: current development and challenges

**Di Wang[1], Simon X. Yang[2]**

[1]School of Information Science and Engineering, Chongqing Jiaotong University, Chongqing 400074, China.
[2]Advanced Robotics and Intelligent Systems (ARIS) Lab School of Engineering, University of Guelph, Guelph ON N1G 2W1, Canada.

**Correspondence to:** Dr. Di Wang, School of Information Science and Engineering, Chongqing Jiaotong University, No. 66, Xuefu Avenue, Nan'an District, Chongqing 400074, China. E-mail: diwang@cqjtu.edu.cn

## Abstract

As a common appearance defect of concrete bridges, cracks are important indices for bridge structure health assessment. Although there has been much research on crack identification, research on the evolution mechanism of bridge cracks is still far from practical applications. In this paper, the state-of-the-art research on intelligent theories and methodologies for intelligent feature extraction, data fusion and crack detection based on data-driven approaches is comprehensively reviewed. The research is discussed from three aspects: the feature extraction level of the multimodal parameters of bridge cracks, the description level and the diagnosis level of the bridge crack damage states. We focus on previous research concerning the quantitative characterization problems of multimodal parameters of bridge cracks and their implementation in crack identification, while highlighting some of their major drawbacks. In addition, the current challenges and potential future research directions are discussed.

**Keywords:** Intelligent detection, crack detection, deep learning, data fusion, feature extraction

## 1. INTRODUCTION

As a pivotal project for economic development and residents' lives and travel, bridges have an irreplaceable role in modern transportation. Bridge structures inevitably incur defects such as pores and cracks when they are affected by overload, temperature change, reinforcement corrosion, construction defects and other factors[1]. If cracks cannot be detected and maintained in a timely manner, traffic safety will gradually be affected, with the potential for bridge collapse and related accidents. Once bridge collapse occurs, it will reduce its structural bearing capacity, which will affect the reliability and safety of the structure, causing immense social repercussions[2-4].

In engineering practice, the health monitoring of bridge cracks has attracted the attention of relevant national departments and research institutions[5,6], corresponding norms and standards have been issued, and the monitoring methods and monitoring parameters for bridge deformation and cracks have been clearly specified.

In recent years, there have been many significant developments in research on the detection of concrete bridge cracks[7-15]. It has been shown that the applications of machine learning in bridge cracks have achieved good results[16-18]. Some studies utilize convolutional neural networks (CNNs) to detect and segment cracks in civil infrastructure with multiple objects[19-23]. It is significant to carry out data-driven research on feature extraction, data fusion and intelligent detection of concrete bridge cracks, which could provide not only a scientific basis for intelligent maintenance of bridges but also data support and a theoretical basis for bridge defect detection, which could have an important role in improving social and economic benefits.

After more than 20 years of scientific research and practice, health monitoring systems have been installed on at least 300 bridges in China. The health monitoring system of a long-span bridge is composed of at least dozens or even hundreds of sensor measuring points. Therefore, numerous monitoring data have been accumulated.

As crack evolution is a gradual and multiscale dynamic process, research is challenging to some extent. There are some common technical difficulties, which will be described from the following three levels:

**(1) Feature extraction of multimodal parameters**
The characterization information related to bridge cracks includes bridge crack shape information, structural mechanics index information, and crack environment information. Bridge crack shape information comprises the multimodal parameters of bridge cracks, such as the length, width and depth of cracks. Structural mechanics index information consists of the dynamic and static elastic modulus, compressive strength, and stress distribution. Crack environment information includes the load, temperature, humidity, and foundation settlement.

From the level of feature extraction for multimodal parameters of bridge cracks, the characterization information of bridge cracks has multiscale and diversity, which makes it difficult to extract multimodal parameter characteristics. The bridge service environment is complex and changeable, and the crack formation mechanism of concrete bridges varies, which generates multiscale and diverse characterization information. The large amount of information hinders feature extraction.

Due to the influence of background interference information, equipment accuracy, data acquisition mode, signal propagation path and propagation medium, there is considerable redundancy in the data, resulting in a weak effective signal in the monitored multimodal data and difficulty in extracting its sensitive features.

**(2) Multisource heterogeneous data fusion representation of bridge cracks**
Different kinds of data, such as the length of a bridge crack, the load, and the environmental humidity, belong to multisource heterogeneous data.

From the level of description for the bridge crack damage state, it is difficult to describe the damage evolution state of bridge cracks due to the large amount of multisource heterogeneous data. When bridge cracks are initially generated, the impact on the bridge is small, and the state change of each indicator is not obvious under the same conditions.

Due to the large variety, quantity, variation in sampling methods and many random interference factors of the sensors, the quality of the acquired data is reduced, and the monitoring data are uncertain and low-density, which makes it difficult for the collected data to accurately and pertinently reflect the evolution of the bridge crack.

**(3) Intelligent detection methods of bridge cracks**
From the level of diagnosis for the bridge crack damage state, the characterization information of bridge cracks is mixed and weak, so it is difficult to scientifically diagnose them. Due to the diversity of monitoring equipment, the diversity of monitoring methods and the variability of monitoring locations, the monitoring data show the characteristics of multimodal, strong correlation and high dimension, which makes the crack information extremely complex, highly mixed and weak separability. It is difficult to analyze the monitoring data and to scientifically describe and diagnose the cracks of concrete bridges. Additionally, the deterioration trend of cracks cannot be scientifically predicted.

By analyzing the feature extraction method of crack multimodal data, the problem of difficult multimodal parameter feature extraction due to the multiscale and diversity of characterization information can be solved. By investigating the multisource heterogeneous data fusion representation of cracks, the difficulty in describing the damage evolution state due to the large amount of multisource heterogeneous data can be addressed. By examining the intelligent prediction model of crack deterioration trends, the problem of difficult scientific diagnosis due to the overlapping and weak separability of bridge crack characterization information can be solved.

The remainder of the article is organized as follows: Section 2 describes the common difficulties in research on feature extraction, data fusion and the detection of concrete bridge cracks. The global research status from three aspects is elaborated and analyzed in Section 3. Section 4 summarizes the principal concluding remarks of the current research with a statement about its strengths and limitations. Figure 1 shows the research framework.

## 2. RESEARCH STATUS
Research on concrete bridges mainly includes three key aspects: feature extraction of multimodal data, fusion representation of multisource heterogeneous data, and intelligent detection models of bridge cracks. The review of the global research status and development trends mainly focuses on these three aspects.

### 2.1. Research status of multimodal data feature extraction of bridge cracks
Although some nondestructive methods, including the ultrasonic pulse[24,25] and elastic wave[26-28], have achieved good results for bridge crack detection, feature extraction from the multimodal data of bridge cracks remains challenging.

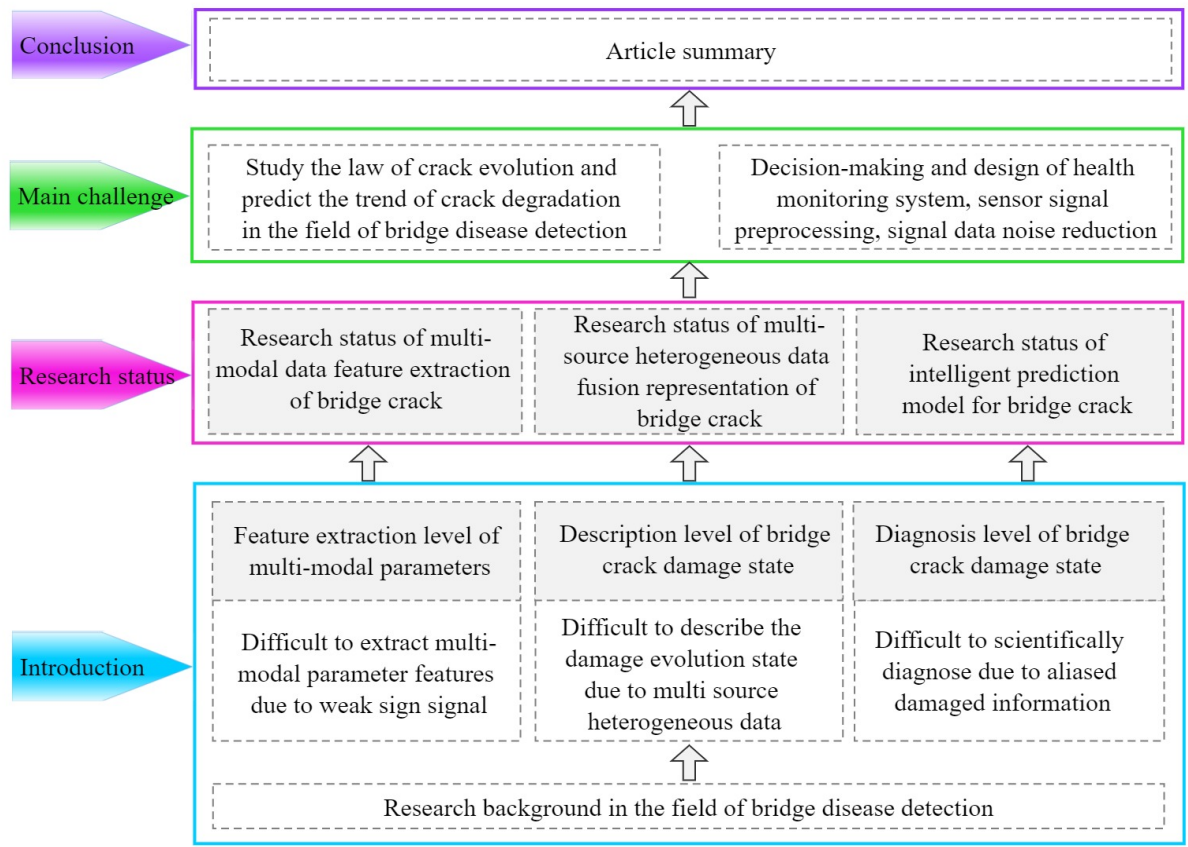

**Figure 1.** Research framework.

The use of intelligent information processing methods to remove background noise, capture multiscale and diverse representation information, and extract its related sensitive features is the basis for detecting concrete bridge cracks.

In the aspect of feature extraction for concrete bridge cracks, the traditional classical methods include principal component analysis (PCA), sparse decomposition, wavelet transform (WT), and adaptive neural fuzzy inference system (ANFIS).

PCA obtains different principal components via the matrix transformation of signals, arranges the sizes of the principal components according to the variance, and compares the contribution rate of each component with the threshold value, thus realizing effective feature extraction[29,30]. In the research on concrete structure crack recognition based on multiple image features, Yu *et al*. collected 1200 bridge crack images and employed integral projection and PCA to extract effective features sensitive to the crack for crack edge detection[31]. Mei *et al*. collected acceleration data from all the vehicles within a certain period and extracted the transformed features that are related to bridge damage with Mel-frequency cepstral coefficients and PCA to identify the damage by comparing the distributions of these transformed features[32]. PCA is a widely utilized method to remove noise and extract effective features, but it cannot accurately analyze the real subspace structure of data.

Sparse decomposition[33] is different from the traditional feature extraction method. Sparse decomposition decomposes noisy signals on redundant dictionaries to achieve feature extraction. In the research of bridge crack identification, Li *et al*. adopted a self-learning algorithm to extract scale invariant features from 27,471 unlabeled bridge pavement crack images, adopted an improved sparse coding representation to obtain a feature dictionary, adopted a spatial pyramid pooling method for feature extraction, and then employed a linear support vector machine (SVM) classifier for crack identification[34]. Wang *et al*. collected hundreds of concrete crack images and proposed a fast detection method for concrete cracks based on L2 sparse representation[35]. To suppress the noise disturbances, discrete cosine transformation is applied to extract the frequency-domain characteristics of the crack and non-crack image regions. The established complete dictionary was used to quickly calculate its sparse coefficient and effectively select candidate cracks via a pooling operation. Sparse decomposition can be succinctly expressed as a linear combination of several bases, more comprehensively and carefully characterize some features covered by a signal and more effectively separate the signal and noise. However, sparse decomposition involves a vast amount of computation and computational complexity.

The WT is mainly based on the different distributions in the frequency domain of the noise and signals and decomposes a noisy signal into multiple scales. Then, the wavelet coefficients belonging to the noise are removed at each scale, while the wavelet coefficients belonging to the signal are retained and enhanced. After wavelet denoising, the signal is reconstructed to achieve effective feature extraction[36]. Nguyen *et al*. measured displacement signals with four separate measurement channels through a sensor system, where the displacement signals were simultaneously transmitted at a sampling speed of 100 samples per second[37]. Then, the WT method was applied to the original deflection signals to decompose the signals into elements and to eliminate interference signals, which extracts the features of multiple cracks under the action of moving loads and improves the sensitivity and accuracy in the identification process. In the study of the finite element model of a simply supported beam with a transverse crack, Nigam *et al*. employed the WT, which uses the deflected edge as input for identifying the crack location in the beam, and obtained a good detection effect[38]. Because WT retains most of the wavelet coefficients of the signal, the image details can be well preserved after noise reduction. As the abrupt part of the signal will not be damaged during noise reduction, it has a good denoising effect. However, since different signals are applicable to different wavelet bases, it is difficult to identify the optimal wavelet base.

The ANFIS[39,40] organically combines fuzzy logic inference and neural networks and performs self-adaptive learning on fuzzy experience and knowledge while applying reasoning similar to the human brain to eliminate noise and interference and to extract feature information[41,42]. Bilir *et al*. applied the results of free shrinkage tests conducted to determine the length changes and ring tests performed to determine the restrained drying shrinkage cracks for predicting the crack widths of granulated blast furnace slag fine aggregate mortars with ANFIS on 456 data and used the replacement ratios, drying time and free shrinkage length changes as inputs and crack width as output to predict the shrinkage cracking of the mortar types[43]. The ANFIS uses the neural network self-learning ability and fuzzy logic reasoning ability to extract fuzzy rules from the dataset and calculates the optimal parameters of the membership function to adaptively mine the sensitive features in the data.

With the rapid development and wide application of artificial intelligence and deep learning, an increasing number of deep learning methods have been applied to concrete bridge crack detection, especially in crack feature extraction[44,45]. In research on the automatic classification of concrete structure crack damage based on cascade generalized neural networks, Guo *et al*. employed 10,000 cracked and uncracked images from wall images and pavement images with a splitting ratio for datasets of 8:2 and proposed a cascade broad neural network for concrete structural crack damage classification, where the multilevel cascade classifier was utilized to extract the characteristics of concrete cracks and achieved an accuracy of 97.9%[15]. Zheng *et al.* adopted a model based on a CNN to amplify and extract the features for 5000 concrete crack images and analyzed the morphological and geometric indices of cracks through the training of building surface data such as roads, bridges, houses and dams, while achieving the highest crack detection accuracy of 98% and the average detection accuracy of 87%[46]. Xu *et al.* proposed an end-to-end crack detection model based on a CNN for 2068 bridge crack images using only images and image labels as input and extracted multiscale crack feature information by using cavity convolution and pooling methods, thus reducing the computational complexity and achieving a high recognition rate (96.37%)[47]. Teng *et al.* applied 11 well-known CNN models as the feature extractor of YOLOv2 for crack detection with 990 RGB bridge crack images, providing a basis for rapid and accurate crack detection of concrete structures, and achieved a high precision of 0.89 and a fast computing speed[48].

The self-attention mechanism is a model in deep learning that has been widely applied in natural language processing tasks in recent years. The idea of attention is to filter out a small amount of important information from a large amount of information and focus on this important information, disregarding most of the unimportant information. The larger the weight, the more focused the corresponding feature, where the weight represents the importance of the feature. The self-attention mechanism reduces the dependence on external information and is better at capturing the internal correlations of data or features.

A schematic diagram of the self-attention mechanism is shown in Figure 2. In the figure, three multimodal features, such as the depth of the crack, the load, and the environmental humidity, are used as input, and the output is their correlation information. The feature vectors of $a^1$, $a^2$ and $a^3$, which have certain meanings, are obtained from the preprocessing of the three multimodal features. They are respectively multiplied by three weight vectors ($w^q$, $w^k$, and $w^v$) to obtain three corresponding vectors ($q^i$, $k^i$, $v^i$, and $i$ denotes the number of features). The following process can be divided into three steps. Firstly, the similarity calculation of $q^i$ and $k^i$ is performed to obtain the weight of feature. The calculation formula is as follows

$$\alpha_{i,j} = q^i k^j \quad i = 1, 2, 3 \tag{1}$$

where $\alpha_{i,j}$ denotes the weight. Then the softmax function is used to normalize the weight of each feature. The normalized weight can be calculated by

$$\hat{\alpha}_{i,j} = \frac{e^{\alpha_{i,j}}}{\sum_{j=1}^{3} e^{\alpha_{i,j}}} \quad i = 1, 2, 3 \tag{2}$$

where $\hat{\alpha}_{i,j}$ denotes the normalized weight. Finally, a weighted sum operation is performed on the normalized weights and each corresponding vector ($V$) to obtain their corresponding outputs. The calculation method is as follows

$$b^i = \sum_{j=1}^{3} \hat{\alpha}_{i,j} v^j \quad i = 1, 2, 3 \tag{3}$$

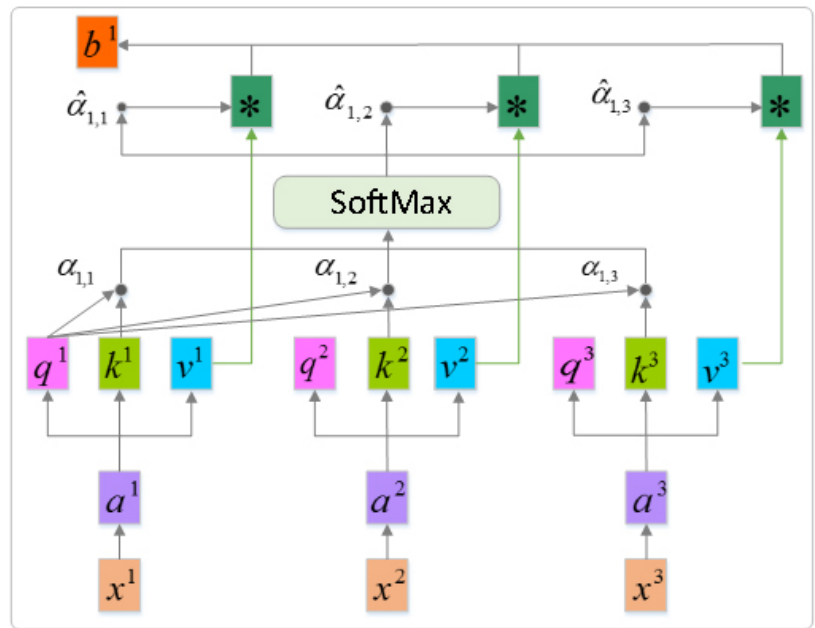

**Figure 2.** Schematic of the self-attention mechanism.

where $b^i$ contains the relevant information among the three input features. Through this way, the self-attention mechanism effectively assigns weight coefficients via the degree of similarity relationship between two feature vectors and quickly extracts relevant information among multimodal parameters.

Pan *et al*. built a spatial-channel hierarchical network with a base net visual geometry Group 19 (VGG19) to automatically detect bridge cracks at the pixel level and applied the self-attention mechanism not only for mining the semantic dependence features of the spatial and channel dimensions but also for adaptively integrating local features into their global dependence features[49]. The segmentation performance of the proposed approach was validated with public datasets containing 11,000 cracked and uncracked images and achieved excellent evaluation results in terms of the mean intersection over union (85.31%). Zhao *et al*. proposed a modified U-net for minute crack segmentation of 200 raw images in real-world, steel-box-girder bridges and applied a self-attention module with softmax and gate operations to obtain the attention vector, which enables the neuron to focus on the most significant receptive fields when processing large-scale feature maps[50]. The self-adaptation module, which consists of a multiplayer perceptron subnet, was selected for deeper feature extraction inside a single neuron. The self-attention mechanism mimics the internal process of biological observation behavior and can quickly extract important features of data, which is especially good at capturing the internal correlation of data or features.

For feature extraction from bridge crack multimodal data, the traditional feature extraction method of representative information has certain limitations, but the self-attention mechanism can reasonably allocate weights among the time domain, spatial domain and channel domain to extract the most relevant features of the target. The methods of feature extraction utilized in the above studies are summarized in Table 1.

**2.2. Research status of the multisource heterogeneous data fusion representation of bridge cracks**
The multisource heterogeneous parameters, such as the operating environment, load and structural mechanical state indices of bridge cracks, have a strong correlation and low density, and it is difficult to accurately and comprehensively reflect the evolution state of cracks. By analyzing and synthesizing the redundant and related monitoring data of the fracture, the data fusion method can more comprehensively and accurately evaluate the evolution state of the fracture. Classical information fusion methods include Bayesian estimation, DS (Dempster-Shafer) evidence theory, and the Kalman filter.

**Table 1. Summarized approaches of feature extraction on bridge cracks**

| Research | Method/approach | Advantages | Drawbacks |
| --- | --- | --- | --- |
| Yu et al.[31]<br>Mei et al.[32] | PCA | Ability to remove noise and extract effective features | Unable to analyze the real subspace structure of data |
| Li et al.[34]<br>Wang et al.[35] | Sparse decomposition | Ability to separate the signal and noise | Mass computation and computational complexity |
| Nguyen et al.[37]<br>Nigam et al.[38] | Wavelet transform | Good denoising effect | Unable to identify the optimal wavelet base |
| Bilir et al.[43] | Adaptive neural fuzzy inference system | Self-learning ability and fuzzy logic reasoning ability | |
| Guo et al.[15]<br>Zheng et al.[46]<br>Xu et al.[47]<br>Teng et al.[48] | Deep learning | Strong expression ability | Long calculation time |
| Pan et al.[49]<br>Zhao et al.[50] | Self-Attention | Ability to quickly extract important features of data | |

Bayesian estimation is based on the prior probability and posterior probability criteria in probability theory and uses conditional probability to represent the uncertain information in the monitoring data[51]. Li *et al*. employed a fully CNN and naive Bayesian data fusion model to automatically segment cracks and noise and fused the extracted multilayer features to obtain significant crack recognition performance[18]. The algorithm was verified with 7200 datasets of bridge substructures collected from 20 in-service bridges under various circumstances. The recognition results showed remarkable performance of the proposed algorithm compared to other recent algorithms. Bayesian estimation must rely on the distribution of the subjective prior probability of the data. However, some of the prior probability distribution of the monitoring data in the actual project is unknown. Therefore, there is a contradiction between this subjectivity and scientific objectivity.

DS evidence theory uses DS synthesis rules to fuse multiple pieces of evidence (feature information), eliminate or reduce the complementarity, redundancy and uncertainty among data, and obtain the fusion judgment or diagnosis of comprehensive features with certain decision rules. Zhao *et al*. applied DS evidence theory to perform weighted fusion on structural health monitoring data from multiple sensors of a two-story concrete frame to provide an accurate and final interpretation of the structural health status[52]. Guo *et al*. applied multiscale space theory and a data fusion method to detect the multiscale damage of beams and plates in a noisy environment and applied DS evidence theory to fuse and express the multiscale damage characteristics in a multiscale space to obtain good anti-noise ability and damage sensitivity[53]. DS evidence theory is mainly utilized to address the reasoning of uncertain information, but it lacks a certain theoretical basis and has potential exponential explosion risk in calculation.

According to the statistical principle, the Kalman filter uses the statistical characteristics of monitoring data and empirical data to perform real-time fusion representation of uncertain and dynamic redundant monitoring data. Prof. Zhang *et al*. employed a Kalman filter to fuse the parameters of the residual generator during the design of two-degrees-of-freedom controllers in a data-driven environment and a residual generator to explain all the stability[54]. Palanisamy *et al*. estimated a Kalman state structure model based on finite element construction, which effectively fuses different types of acceleration, strain and tilt response data, minimizes internal measurement noise, and realizes the overall response measurement of bridge crack structures[55]. When the state of the monitoring system satisfies the Gaussian distribution and the linear model, the Kalman filter can perform a good fusion representation of the uncertain information and realize the optimal estimation. However, when the system does not satisfy the conditions of the Gaussian distribution and the linear model, this method will be limited.

In recent years, deep neural networks have become a controversial topic for scholars, and their application in data fusion has also achieved good results. Professor Li Hui *et al.* applied a fusion CNN for crack identification from real-world images containing complicated disturbance information (cracks, handwriting scripts, and background) inside steel box girders of bridges with 350 raw images as input[56]. The results demonstrated that the recognition errors of the fusion CNN in both the training and validation processes are smaller than those of the regular CNN. Li *et al.* developed a flexible crack recognition system for the complex bridge crack detection environment, which uses sliding window technology to process the acquired images and uses a trainable context encoder network to fuse the crack image information and features to achieve pixel-level bridge crack detection[57]. Chen *et al.* proposed a deep learning framework based on a CNN and naive Bayesian data fusion scheme to analyze the crack detection of a single video frame and proposed a novel data fusion scheme to extract spatiotemporal coherence information of cracks in videos to improve the overall performance and robustness of the system[58]. Yang *et al.* proposed an effective concrete crack segmentation network based on UAV-enabled edge computing and applied the method of atrous spatial pyramid pooling to realize free multiscale feature extraction and to fuse different levels of feature map information into lower-level features for crack detection[59].

Granular computing realizes the fusion representation of incomplete, imprecise and multimodal information by selecting the granularity space suitable for the diagnosis problem and via granularity transformation in different partition spaces. Granular computing[60] is a natural model that simulates human thinking and the solving of problems. It is a new paradigm for data analysis and problem-solving based on the relationship between two particles. Granular computing provides multi-granularity, multiperspective and multilevel description, reasoning and solution strategies for the satisfactory solution of complex problems and effectively fuses and represents massive multimodal data.

For multisource heterogeneous data fusion representations of fractures, although many traditional data fusion algorithms can fuse and represent multisource heterogeneous data, the monitored data have high dimensions and strong correlations and comprise spatiotemporal data. Traditional information fusion methods have difficulty mining the internal deep and complex feature information. The granularity space in granular computing is consistent with the multilevel and space-time characteristics of bridge health monitoring data, and these monitoring data can be fused and represented according to the granularity space that matches the crack damage diagnosis problem of the bridge.

The abovementioned methods for multisource heterogeneous data fusion representation are summarized in Table 2.

### 2.3. Research status of intelligent detection models for bridge cracks

The monitoring data is mixed, and the fracture evolution state and characterization information present a nonlinear relationship. The separability of this characterization information is weak, so it is difficult to analyze the monitoring data, accurately evaluate the fracture state and predict the development trend. Presently, intelligent detection methods can be divided into shallow network models, wide learning network

Table 2. Multisource heterogeneous data fusion representation methods

| Research | Method/approach | Advantage | Drawback |
|---|---|---|---|
| Li *et al*.[18] | Bayesian estimation | Simple operation | Reliance on the distribution of the subjective prior probability of data |
| Zhao *et al*.[52]<br>Guo *et al*.[53] | DS evidence theory | Applies reasoning to address uncertain information | Exponential explosion risk |
| Zhang *et al*.[54]<br>Palanisamy *et al*.[55] | Kalman filter | Sufficient processing of data subject to Gaussian distribution | Inadequate processing of non-Gaussian data |
| Li *et al*.[56]<br>Li *et al*.[57]<br>Chen *et al*.[58]<br>Yang *et al*.[59] | Deep learning | Strong expression ability | Long calculation time |
| Li *et al*.[60] | Granular computing | Sufficient processing of multi-granularity spatiotemporal data | |

models and deep network models.

The shallow network model adopts a neural network with one single hidden layer and mines the characteristic information in the nonlinear data to realize the intelligent identification and diagnosis of scientific problems. Classic shallow diagnosis algorithms mainly include extreme learning machines, BP neural networks, and SVMs. Wang *et al*. proposed a multiview multitask crack detector to calculate various visual features (such as texture and edge) of the image area, suppress various background noise, and emphasize the separability of crack region features and complex background features[61]. The experimental results with 350 crack images showed that the proposed method improves the training efficiency with a precision of 92.3% and a recall of 89.7%. Yan *et al*. used a simple supported beam with a single crack and double cracks as an example to identify local cracks and proposed a damage identification method for beam structures based on a BP neural network and SVM[62]. The results showed that the strain mode difference curve at the damaged part undergoes considerable changes, and better identification accuracy is obtained, where the recognition efficiency for a single crack is 99.9% and 100% for a double crack with the BP neural network and 99.6% for a single crack and 99.9% for a double crack with the SVM. Liu *et al*. selected the cantilever beam as the research object and proposed a crack damage detection method with a BP neural network based on the curvature modulus ratio, the natural frequency, which uses the parameters of frequency relative attenuation and the first-order maximum curvature modulus ratio as the input, and the parameters of crack position and damage degree as the output and achieves a good crack identification effect in terms of a relative error of 1.7 on crack images with a size of 248.3 mm and an injury degree of 84.6%[63]. The shallow intelligent diagnosis algorithm has a hidden layer, which can mine complex feature information and be applied for intelligent diagnosis in many fields. However, when the data present massive growth and have multilevel high-dimensional features, the shallow intelligent diagnosis method is limited.

The broad learning system (BLS) method[64,65] is a kind of neural network that does not depend on the structure depth and can realize the diagnosis of research problems by widening the width of the network for information mining. Guo *et al*. adopted a network structure that combines deep learning and BLS on 40,000 concrete crack images, intelligently trains the network with the original image via linear and nonlinear mapping processes with dynamic updating of the weights, and performs binary classification of concrete surface cracks[66]. The results proved that the accuracy of the presented method achieves 98.55% and 96.12% and that the training time is 59.8 s and 95.76 s with two different datasets. Chen *et al*. and Xu *et al*. proposed a recursive BLS, which uses recursive connections at the enhancement nodes of the network to capture the dynamic characteristics of time series and shows excellent performance on the chaotic time series[67,68]. Their experimental results showed that the accuracy achieved 91.96% with a training time of 123 minutes with the MS-Celeb-1 M face database and that the RMSE was reduced to 121.9 on air quality dataset prediction. The BLS performs abstract representation and incremental learning of dynamic high-dimensional data and achieves a high diagnostic accuracy, short time and strong real-time performance. However, the BLS is inferior to deep networks in terms of the deep data mining performance of feature extraction and has a poor effect on detection problems for time-dependent data.

The deep neural network model uses not only a deep neural network to represent the multilayer features of the identified object but also the extracted high-level features to reflect the intrinsic nature of the data, which has better robustness and diagnostic ability than the shallow network. Classic deep intelligent diagnosis methods include CNNs, deep belief networks and recursive neural networks. Li *et al.* employed the sliding window algorithm to divide not only the bridge crack image into slices with a size of 16 pixels × 16 pixels but also the slices into the bridge crack surface element and bridge background surface element[69]. The authors then proposed a deep bridge crack classification model based on CNN for the identification of bridge background surface elements and bridge crack surface elements. The proposed algorithm achieved an average accuracy of 94.5% with 2000 bridge crack images. Islam *et al.* established a deep CNN using an encoder and decoder framework for semantic segmentation to realize pixel-level automatic detection of bridge cracks, obtaining scores of approximately 92% for both the recall and F1 value[70]. Liang *et al.* utilized a double CNN model to identify cracks in actual concrete bridges, which highly improved the reliability and accuracy of identification with accuracies of 98.6% and 99.5% by the CNN and FCN models, respectively[71]. Li *et al.* proposed a new type of fully connected state, long-term and short-term memory neural network, which was employed to discriminate sensor faults and structural damage without knowing the fault details, and obtained excellent performance with an RMSE of 0.03[72]. The deep network model has made progress in big data processing. Although the training takes a long time due to the numerous network layers, complex structure, and many super parameters, it can process the data with complex internal information characteristics and is more suitable for the engineering needs of timely warning for dangerous situations in bridge health monitoring.

For intelligent crack detection model construction, a shallow network model can mine complex data for bridge crack detection, but its feature extraction ability is weak, and it cannot carry out multilayer feature representation for recognized objects. Thus the recognition rate is low and the generalization performance is mediocre. The BLS can mine and integrate multilevel and multimodal, complex state information by widening the network and incremental learning. The BLS achieves high diagnostic accuracy, fast speed and strong real-time performance for the research target but has a poor effect on the time-dependent data detection problem. The deep network model can represent the multilayer features of the recognized object and use the extracted high-level features to reflect the intrinsic nature of the data. The model has the ability of deep information mining and achieves better robustness, generalization and recognition performance.

The structure of the deep neural network mapping model is shown in Figure 3, and the methods utilized for intelligent detection in the above review articles are summarized in Table 3.

## 3. MAIN CHALLENGES
Despite great developments in the research on concrete bridge crack identification based on data, certain challenges exist.

As a cavity, the crack has three-dimensional information with length, width and depth. However, traditional crack identification is usually based on the recognition with the two-dimensional surface information (length and width) of cracks, and research on three-dimensional cracks is limited. Although three-dimensional crack reconstruction technology has gradually attracted researchers' attention[73,74], its evolution mechanism remains ambiguous. There is an urgent need in the field of bridge disease detection to investigate the law of crack evolution and to predict the trend of crack degradation.

**Table 3. Intelligent detection methods**

| Research | Method/approach | Advantage | Drawback |
| --- | --- | --- | --- |
| Wang et al.[61]<br>Yan et al.[62]<br>Liu et al.[63] | Machine learning | Excellent predictive performance | Unsuitable for large sample data |
| Guo et al.[66]<br>Xu et al.[67,68] | Broad learning system | High accuracy and short time | Unsuitable for sequential data |
| Li et al.[69]<br>Islam et al.[70]<br>Liang et al.[71]<br>Li et al.[72] | Deep learning | Sufficient for deep mining big data information | Long computing time |

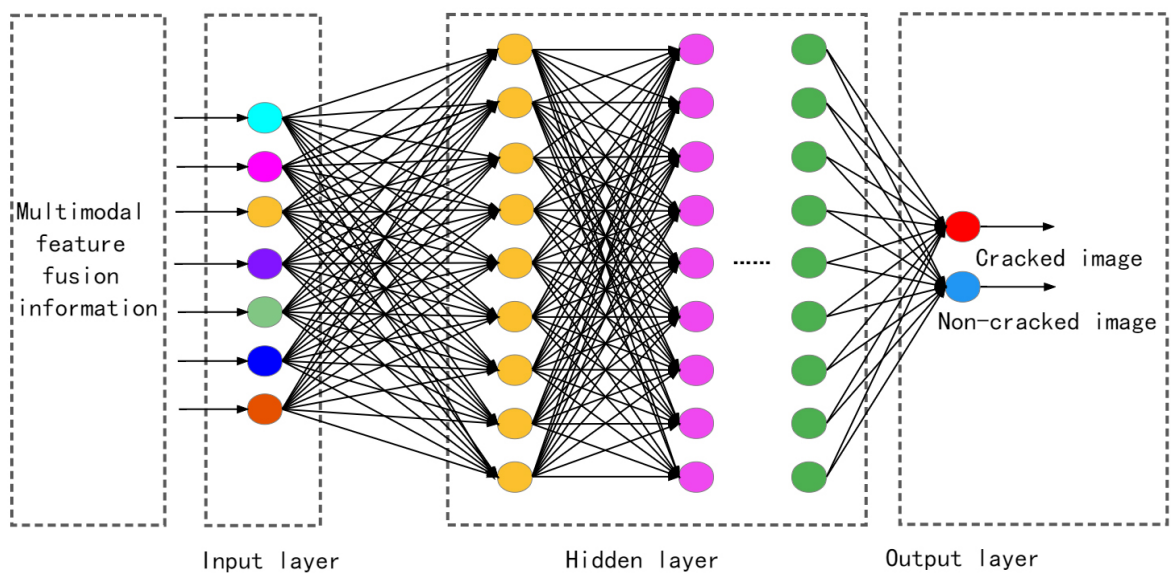

**Figure 3.** Structure of the deep neural network mapping model.

In bridge crack research based on data, bridge health monitoring systems has an irreplaceable role. The main purpose of research on the long-span bridge health monitoring system is to accumulate the design and scientific research data of bridge health monitoring, realize the real-time damage diagnosis and safety assessment of the structure, and support management and maintenance decision-making. Presently, the challenges faced in bridge health monitoring mainly include decision-making and the design of health monitoring systems, sensor signal preprocessing, and signal data noise reduction.

## 4. CONCLUSION

This paper presents a comprehensive review of recent advances in the field of data-driven bridge crack health detection. The latest achievements in bridge crack feature extraction, data fusion and intelligent detection are introduced. Based on the discussion of the three technical difficulties of bridge crack multimodal data feature extraction, multisource heterogeneous data fusion representation, and intelligent crack detection model construction, the latest progress in bridge crack detection research is summarized in detail, and their major advantages and drawbacks in this field are highlighted. The main current challenges and potential future research directions are also discussed.

## DECLARATIONS

### Authors' contributions

Made substantial contributions to the research and investigation process, reviewed and summarized the literature, wrote and edited the original draft: Wang D

Performed oversight and leadership responsibility for the research activity planning and execution, as well as performed critical review, commentary and revision: Yang SX

### Availability of data and materials
Not applicable.

### Financial support and sponsorship
This work was supported by the National Natural Science Foundation of China (Grant No. 62103068; Grant No. 51978111), Science and Technology Research Project of Chongqing Education Commission (Contract No. KJQN202100745).

### Conflicts of interest
All authors declared that there are no conflicts of interest.

### Ethical approval and consent to participate
Not applicable.

### Consent for publication
Not applicable.

### Copyright